\documentclass{article} 
\usepackage{iclr2026_conference,times}

\usepackage{amsmath,amsfonts,bm}

\def\eqref#1{equation~\ref{#1}}

\def\1{\bm{1}}

\def\vf{{\bm{f}}}

\def\vx{{\bm{x}}}

\def\mX{{\bm{X}}}

\DeclareMathAlphabet{\mathsfit}{\encodingdefault}{\sfdefault}{m}{sl}
\SetMathAlphabet{\mathsfit}{bold}{\encodingdefault}{\sfdefault}{bx}{n}

\def\gD{{\mathcal{D}}}

\def\gG{{\mathcal{G}}}

\def\gL{{\mathcal{L}}}

\def\gN{{\mathcal{N}}}

\newcommand{\E}{\mathbb{E}}

\newcommand{\R}{\mathbb{R}}

\usepackage{xcolor}%
\definecolor{citecolor}{HTML}{0071bc}
\usepackage[colorlinks=true, allcolors=citecolor]{hyperref}
\usepackage{url}
\usepackage{caption}
\usepackage{graphicx}
\usepackage{booktabs}
\usepackage{multirow}
\usepackage{makecell}
\usepackage{wrapfig}
\usepackage{siunitx}
\usepackage{bbding,pifont}
\usepackage{amsmath,amssymb,amsfonts,amsthm,mathrsfs}

\usepackage{algorithm}
\usepackage{algpseudocode}

\usepackage[title]{appendix}

\title{Gait Recognition via Collaborating Discriminative and Generative Diffusion Models}

\author{Haijun Xiong \quad Bin Feng \footnotemark[2] \quad Bang Wang \quad Xinggang Wang \quad Wenyu Liu \\ \\
School of EIC, Huazhong University of Science \& Technology \\
\texttt{\{xionghj,fengbin\}@hust.edu.cn}
}

\newcommand{\ie}{\textit{i}.\textit{e}.}
\newcommand{\eg}{\textit{e}.\textit{g}.}

\newcommand{\name}{CoD\textsuperscript{2}}
\newcommand{\highcontrol}{High-level Control Module}
\newcommand{\multicondition}{Multi-level Conditional Control}
\newcommand{\normaldistribution}{\mathcal{N}(0, I)}

\iclrfinalcopy
\begin{document}

\maketitle

\begin{abstract}
Gait recognition offers a non-intrusive biometric solution by identifying individuals through their walking patterns. Although discriminative models have achieved notable success in this domain, the full potential of generative models remains largely underexplored. In this paper, we introduce \textbf{\name{}}, a novel framework that combines the data distribution modeling capabilities of diffusion models with the semantic representation learning strengths of discriminative models to extract robust gait features. We propose a \multicondition{} strategy that incorporates both high-level identity-aware semantic conditions and low-level visual details. Specifically, the high-level condition, extracted by the discriminative extractor, guides the generation of identity-consistent gait sequences, whereas low-level visual details, such as appearance and motion, are preserved to enhance consistency. Furthermore, the generated sequences facilitate the discriminative extractor's learning, enabling it to capture more comprehensive high-level semantic features. Extensive experiments on four datasets (SUSTech1K, CCPG, GREW, and Gait3D) demonstrate that \name{} achieves state-of-the-art performance and can be seamlessly integrated with existing discriminative methods, yielding consistent improvements.

\end{abstract}

\renewcommand{\thefootnote}{\fnsymbol{footnote}}
\footnotetext[2]{Corresponding author.}

\section{Introduction}

\begin{figure}[t]
    \centering
    \includegraphics[width=0.90\linewidth]{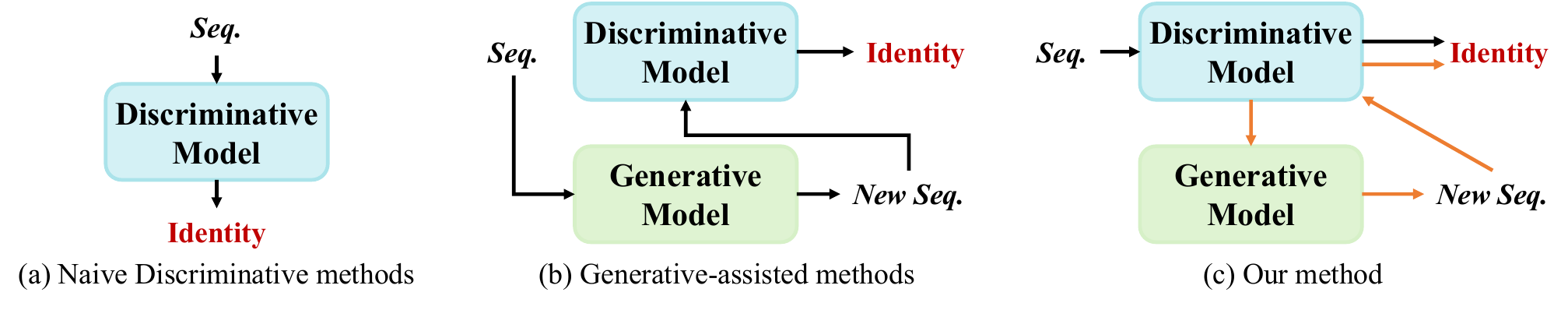}
    \vspace{-10pt}
    \caption{\textbf{Comparison of different methods for gait recognition.} (a) Naive discriminative methods, such as GaitSet~\citep{GaitSet}; (b) Generative-assisted methods, such as DenoisingGait~\citep{DenoisingGait}; (c) Our proposed \name{}, which integrates collaborating discriminative and generative models.}
    \vspace{-15pt}
    \label{fig: motivation}
\end{figure}

Gait recognition is a biometric technology that distinguishes individuals based on unique walking patterns. Unlike other biometric modalities, such as face, iris, and fingerprint recognition, gait can be captured from a distance without requiring subject cooperation, making it particularly suitable for applications in crime prevention, sports science, and healthcare~\citep{venkat2011robust, sepas2022deep}. Despite significant progress in gait recognition, existing discriminative methods~\citep{DyGait, BigGait, MambaGait} (\autoref{fig: motivation}~(a)) continue to struggle in complex scenarios involving variations in clothing, viewpoints, occlusions, and carried objects, which complicate the extraction of robust discriminative features.

Generative models, particularly diffusion models~\citep{DDPM, DDIM}, have recently gained significant attention for their remarkable capability to generate high-quality images, visually compelling images. These models excel at capturing complex data distributions and generate realistic samples by iteratively reversing a noise injection process. Beyond image synthesis, the potential of diffusion models has been increasingly explored in video generation~\citep{VideoDM}, where they effectively capture temporal coherence and high-level structural dynamics. Such characteristics make them especially suitable for tasks that demand both realistic visual generation and consistent motion evolution, including video synthesis and dynamic scene modeling~\citep{4Real, CAT4D}. Furthermore, due to their powerful representational capacity, recent works have leveraged pre-trained diffusion models for a variety of downstream applications, achieving promising results in pose estimation~\citep{DiffPose}, mesh recovery~\citep{DPMesh, HMDiff}, and action recognition~\citep{MacDiff, DD-GCN}.

Previous studies~\citep{DenoisingGait} (\autoref{fig: motivation}~(b)) have employed diffusion models to denoise RGB gait sequences and generate clean gait representations. However, such methods do not fully exploit the intrinsic relationship between generative and discriminative models, thereby limiting the potential of the generative model. While discriminative models emphasize inter-class separability, generative models focus on modeling the underlying data distribution. These two paradigms provide complementary perspectives on the data, and their integration can yield a more holistic understanding of gait patterns. Consequently, combining discriminative models with generative diffusion models is essential for enhancing the feature extraction capabilities of both, ultimately leading to more effective gait recognition.

To address these aforementioned challenges, we propose a novel gait recognition framework, \textbf{\name{}}. As illustrated in \autoref{fig: motivation}~(c), \name{} differs fundamentally from prior works by integrating the data distribution modeling capability of diffusion models with the semantic representation learning strength of discriminative models, thereby extracting more robust gait features. We further present a \multicondition{} strategy that combines both high-level and low-level conditions to guide the generative learning process of the diffusion model. Specifically, the high-level condition, derived from the discriminative feature extractor, provides identity-aware semantic information to generate identity-consistent gait sequences. In contrast, the low-level condition preserves essential visual details, such as appearance and motion information, which are critical for maintaining identity consistency in the generated sequences. Moreover, the generated sequences in turn promote the training of the discriminative extractor, enabling it to capture richer and more comprehensive semantic representations. We evaluate \name{} through extensive experiments on four datasets~\citep{SUSTech1K, CCPG, GREW, Gait3D}, achieving state-of-the-art Rank-1 performance. Furthermore, integrating \name{} with four representative discriminative methods~\citep{GaitSet, GaitGL, GaitBase, DeepGaitV2} consistently improves performance across all datasets, demonstrating its strong versatility. Notably, \name{} introduces only a marginal increase in training consumption, with no impact on testing efficiency. In summary, the main contributions are as follows:
\begin{itemize}
    \item We introduce \name{}, a novel gait recognition framework that integrates the data distribution modeling capacity of generative diffusion models with the semantic representation learning ability of discriminative models, enhancing gait feature extraction through their complementary strengths.
    \item We propose a \multicondition{} strategy that jointly leverages high-level identity-aware semantic features with low-level visual details to guide the diffusion model's generative process. The generated sequences facilitate the discriminative model’s learning, further improving feature robustness.
    \item Extensive experiments demonstrate that \name{} achieves state-of-the-art performance and can be seamlessly integrated with existing discriminative methods, consistently improving performance with minimal impact on training consumption and no effect on testing efficiency.
\end{itemize}

\section{Related Work}
\subsection{Gait Recognition}
Current gait recognition methods can be broadly categorized into model-based and appearance-based methods, depending on the input modality.

Model-based methods~\citep{GaitGraph, GaitGraph2, CycleGait, GPGait} exploit structural human priors, such as skeletons and 3D meshes. For example, PoseGait~\citep{PoseGait} integrates multiple skeleton-based features with human prior knowledge to enhance recognition performance, while CAG~\citep{CAG} employs adaptive conditional networks to extract fine-grained representations. Other studies~\citep{GaitMixer, GaitTR} adopt transformer architectures to capture long-range spatial dependencies, and SMPLGait~\citep{Gait3D} further improves recognition by utilizing dense 3D mesh representations reconstructed from RGB images.

Appearance-based methods~\citep{GaitBase, QAGait, DANet, GLGait, MTSGait, HSTL, ParsingGait, GaitGS, XGait} directly learn spatial-temporal representations from gait silhouettes or RGB sequences. GaitSet~\citep{GaitSet} is the first to treat gait sequences as unordered frame sets. Subsequent methods~\citep{GaitPart, CSTL, GaitGL} adopt 1D or 3D CNNs to model local motion patterns across frames, while deeper architectures~\citep{VPNet, DeepGaitV2} have been developed to extract richer identity-discriminative features. Recent studies~\citep{GaitGCI, GaitCSV, CLTD} revisit gait recognition from a causal inference perspective, and DenoisingGait~\citep{DenoisingGait} employs diffusion models to generate noise-free gait representations. Moreover, alternative modalities, such as point clouds and RGB videos, have recently been incorporated into gait recognition frameworks~\citep{SUSTech1K, BigGait}, broadening the scope of this research field.

\subsection{Diffusion Models for Representation Learning}
Diffusion models have emerged as a powerful paradigm for generative modeling, particularly in image and video synthesis~\citep{DDPM, VideoDM}. These models generate high-quality visual content by progressively refining Gaussian noise through an iterative denoising process. Building on their remarkable success, recent studies have extended diffusion models to a wide range of downstream tasks~\citep{FinePose, MacDiff, DiffPose, HumanMac, P2P-Bridge, SatSynth, RAVE, FreeDiff}. For example, DPMesh~\citep{DPMesh} leverages spatial structural priors from pre-trained diffusion models to reconstruct occluded human meshes, while HOIAnimator~\citep{HOIAnimator} introduces Perceptive Diffusion Models to enhance the realism of human–object interactions in animations. Moreover, ControlNet~\citep{ControlNet} integrates spatial conditioning mechanisms into pre-trained diffusion models for precise detail manipulation, and AYG~\citep{AYG} combines Gaussian Splatting with diffusion models to enable text-to-4D generation.

In this paper, we propose \name{}, the first framework that enhances feature extraction by unifying the semantic representation learning capability of discriminative models and the data distribution modeling power of generative models.

\section{Methodology}
\subsection{Background}
Before introducing our proposed method, we briefly review the key concepts of gait recognition and the Denoising Diffusion Probabilistic Model (DDPM)~\citep{DDPM}.

\begin{figure}[!t]
    \centering
    \includegraphics[width=\linewidth]{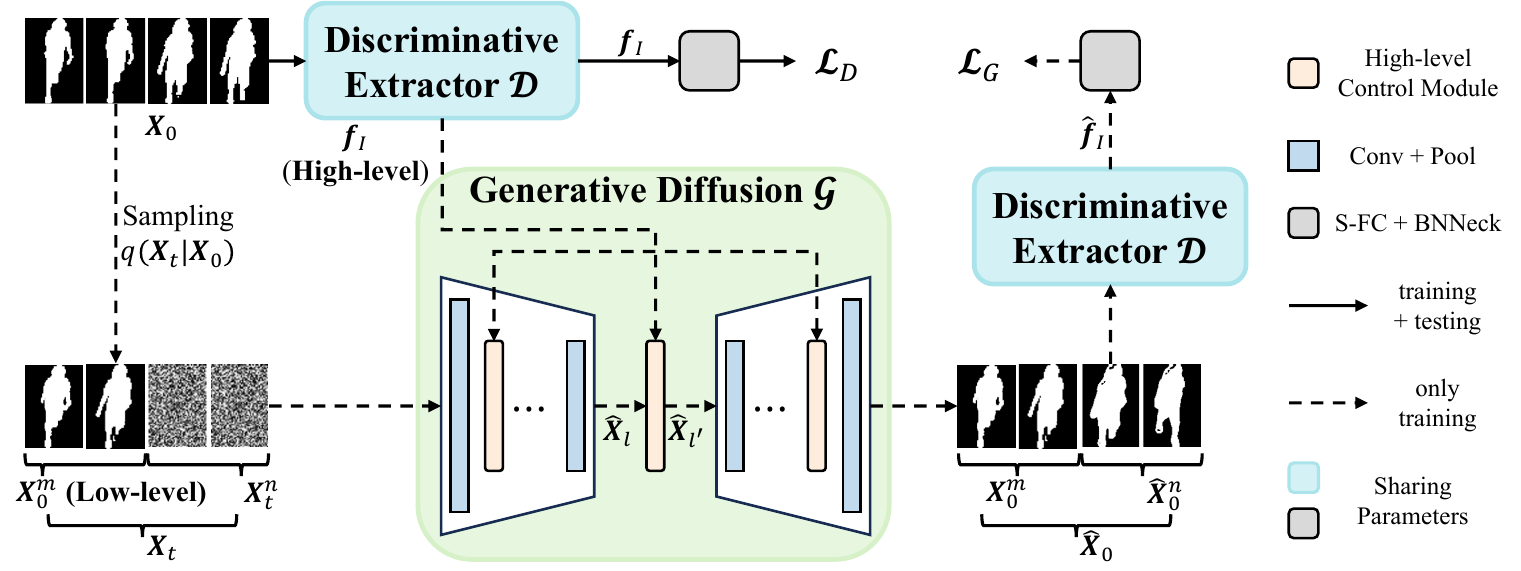}
    \vspace{-10pt}
    \caption{\textbf{Overview of our proposed method.} The discriminative extractor $\gD$ (\eg, GaitSet, GaitGL, GaitBase, or DeepGaitV2) first extracts the identity feature $\vf_I$ from the input gait sequence $\mX_0$. This feature serves as a high-level semantic condition to guide the generative diffusion model $\gG$ during sequence generation. The noise sequence $\mX_t$ is composed of Gaussian noise $\mX_t^n \sim \normaldistribution$ and low-level visual information $\mX_0^m$ sampled from $\mX_0$. The generated gait sequence $\hat{\mX}_0$ is then processed by $\gD$ to extract its identity feature $\hat{\vf}_I$. Finally, $\gD$ and $\gG$ are jointly optimized with the loss $\gL_D$ and $\gL_G$, where $\gG$ is employed only for training, while $\gD$ is used for both training and inference.}
    \vspace{-15pt}
    \label{fig: framework}
\end{figure}

\textbf{Discriminative Gait Recognition.} Given a gait sequence $\mX_0 \in \R^{1 \times T \times H \times W}$ with $T$ frames, each of size $(H, W)$, discriminative gait recognition methods typically process $\mX_0$ through a discriminative feature extractor $\gD$ to obtain the identity representation $\vf_I \in \R^{C \times p}$, where $C$ and $p$ denote the number of channels and parts, respectively:
\begin{equation}
    \vf_I = \gD(\mX_0).
\end{equation}
Subsequently, $\vf_I$ is refined using a separate fully connected (S-FC) layer followed by BNNeck, and optimized with a combination of triplet and cross-entropy losses:
\begin{equation}
    \gL_D = \gL_{tri} + \gL_{ce}.
    \label{equ: discri-loss}
\end{equation}

\textbf{DDPM.} DDPM generates high-quality visual content by iteratively denoising random Gaussian noise. It consists of two phases: a fixed forward diffusion process and a learnable reverse denoising process. In the forward phase, Gaussian noise is gradually added to the original image $x_0$ through a Markov chain, progressively transforming it into pure Gaussian noise $x_T \sim \normaldistribution$. At each timestep $t$, the noised variable $x_t$ depends only on its previous state $x_{t-1}$, as formulated by:
\begin{equation}
    q(x_t|x_{t-1}) = \gN(x_t; \sqrt{1-\beta_t} x_{t-1}, \beta_t I),
\end{equation}
where $\beta_t$ denotes a predefined variance schedule. The reverse process reconstructs $x_0$ from $x_T$ through iterative denoising:
\begin{equation}
    p_\theta(x_{t-1}|x_t) = \gN(x_{t-1}; \mu_\theta(x_t, t), \sigma_t^2 I),
\end{equation}
where $\mu_\theta(x_t, t)$ is a parameterized function, typically implemented as a neural network, used to predict the mean $\hat{\mu}$ at each timestep. Recent methods, such as ControlNet~\citep{ControlNet}, extend diffusion models to controllable generation by incorporating conditional input. Given a condition $c$, the training objective can be formulated as:
\begin{equation}
    \min_\theta \E_{x_0, c, t, \mu} \left[\left \Vert \mu - \mu_\theta(x_t, c, t)\right\Vert_2^2\right],
\end{equation}
which enables the generation of realistic samples from Gaussian noise.

\subsection{Pipeline}
The overall framework of \name{} is illustrated in \autoref{fig: framework}. It comprises two discriminative extractors with shared parameters, and a generative diffusion module. Similar to previous methods, the first discriminative extractor $\gD$ processes the input gait sequence $\mX_0$ to obtain the identity feature $\vf_I$. Meanwhile, a noise sequence $\mX_t$ is constructed by combining the low-level condition $\mX_0^m$ (a part of $\mX_0$) with Gaussian noise $\mX_t^n$. The identity feature, serving as a high-level condition, guides the denoising process of the generative diffusion module $\gG$ by embedding identity-aware semantic information, resulting in a generated gait sequence $\hat{\mX}_0$. The second extractor $\gD$ is then reapplied to extract the identity feature $\hat{\vf}_I$ from $\hat{\mX}_0$, ensuring identity consistency. This bidirectional interaction between $\gD$ and $\gG$ not only reinforces the generative module but also enhances the discriminative extractor’s ability to capture more effective gait features.

\subsection{Discriminative extractor and Generative diffusion module}
The discriminative extractor $\gD$ serves as the core backbone and can be instantiated with various existing gait recognition models, such as GaitSet, GaitGL, GaitBase, and DeepGaitV2-P3D (abbreviated as DeepGaitV2). The versatility of our method is further validated in \autoref{tab: versatility}. Considering that binary silhouette sequences are substantially simpler than RGB inputs and that directly predicting noise from noisy sequences provides limited discriminative information~\citep{MacDiff, I2v-adapter}, we adopt a lightweight generative diffusion module $\gG$ to generate new sequences from noise. The architectural details of $\gG$ are presented in \autoref{sec:appendixa}.

\subsection{Multi-level Conditional Control}
The generative diffusion module $\gG$ takes the noise sequence $\mX_t$ and the identity feature $\vf_I$ as input. Here, $\vf_I$ serves as a high-level control condition, encapsulating identity-aware semantic information. Meanwhile, $\mX_0^m$ in $\mX_t$, derived from the original sequence $\mX_0$, preserves low-level visual cues (such as appearance and motion), acting as a low-level control condition during the denoising process.

\textbf{Low-level conditional control.} Unlike text-to-video generation, gait sequence generation requires preserving visual details from original sequences, such as appearance and motion information. To achieve this, we introduce a sampling strategy that randomly selects continuous $m$ frames from $\mX_0$ as a reference, denoted as $\mX_0^m \in \R^{1\times m \times H \times W}$. This reference is concatenated with Gaussian noise $\mX_t^n \in \R^{1 \times (T-m) \times H \times W}$ along the temporal dimension to construct the noise sequence $\mX_t \in \R^{1\times T \times H \times W}$, formulated as:
\begin{equation}
    \begin{split}
        \mX_0^m &= \mX_0[k:k+m], k\in [0, T-m], \\
        \mX_t^n &= \text{Sample}(\normaldistribution), \\
        \mX_t &= \text{Cat}(\mX_0^m, \mX_t^n),
    \end{split}
    \label{equ: sample}
\end{equation}
where $\text{Cat}(\cdot)$ denotes the concatenation operation. During denoising, the spatial-temporal modeling process transfers low-level visual cues from $\mX_0^m$ to $\mX_t^n$, ensuring that the generated sequences retain essential appearance and motion details. Inspired by LAMP~\citep{LAMP}, we keep the reference frames $\mX_0^m$ noise-free during training, meaning that $\mX_0^m$ remains unchanged after passing through a 3D convolutional layer in $\gG$, \ie, 
\[
\left[\hat{\mX}_0^m,\hat{\mX}_i^n\right]= \text{Conv}\left(\left[\mX_0^m,\mX_i^n\right]\right), \quad \hat{\mX}_0^m = \mX_0^m,
\]
which preserves both temporal identity consistency and the integrity of low-level visual details during denoising. By preserving low-level details, this strategy enhances control effectiveness and improves the overall sequence generation.

\begin{wrapfigure}[13]{r}{0.5\textwidth}
\vspace{-6pt}
\centering
\includegraphics[width=\linewidth]{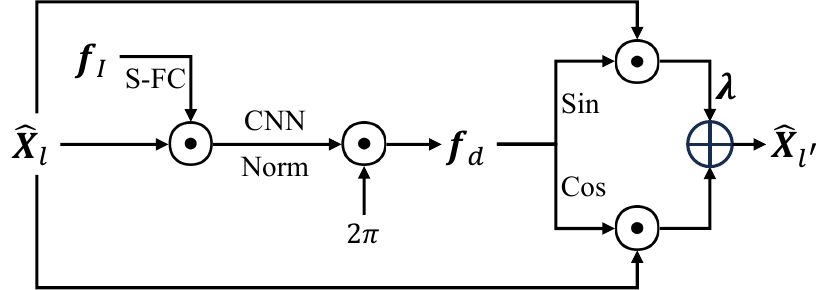}
\vspace{-15pt}
\caption{\textbf{Details of \highcontrol{}}. The S-FC denotes a separate fully connected layer, and $\boldsymbol{\lambda} \in \R^{C'}$ is a learnable channel-wise control vector that regulates the adjustment intensity across different feature channels.}
\label{fig: high-level}
\end{wrapfigure}

\textbf{High-level conditional control.} The high-level condition embeds identity-aware semantic information into the generative diffusion module $\gG$, providing effective guidance during the generation process. While ControlNet~\citep{ControlNet} performs element-wise addition for condition fusion after convolutional layers, we find this operation too coarse for gait sequence generation, leading to degraded performance (as shown in \autoref{tab: ablation_highlevel}). To address this limitation, we propose a refined \highcontrol{} that seamlessly integrates $\vf_I$ into $\gG$, facilitating identity-aware guidance and improving the generated sequences of generated sequences (as illustrated in \autoref{fig: high-level}).

We draw inspiration from Euler's formula:
\begin{equation}
e^{i\theta} = \cos(\theta) + i \sin(\theta),
\label{equ: euler}
\end{equation}
which represents a signal as a rotation in the complex plane, thereby encoding both amplitude and phase information. Motivated by this, we design a phase modulation module based on sinusoidal projection to effectively embed high-level identity semantics into the generative process.

Specifically, given an intermediate noisy sequence $\hat{\mX}_l \in \mathbb{R}^{C' \times T \times H \times W}$ from the reverse diffusion process, we compute a spatially varying phase feature $\vf_d$ conditioned on the identity feature $\vf_I$:
\begin{equation}
    \vf_d = 2\pi \cdot \text{Norm}(\text{Conv}(\hat{\mX}_l \cdot \text{S-Fc}(\vf_I))).
\end{equation}
Here, $\text{S-FC}(\vf_I)$ denotes a spatially broadcasted identity embedding, and $\text{Norm}(\vx) = \frac{\vx - \vx_{\text{min}}}{\vx_{\text{max}} - \vx_{\text{min}}}$ normalizes the values to the range $[0, 2\pi]$ via min-max normalization. We then apply sinusoidal modulation to inject identity-aware semantics into the sequence:
\begin{equation}
    \hat{\mX}_{l'} = \hat{\mX}_l \cdot \cos(\vf_d) + \boldsymbol{\lambda} \cdot \hat{\mX}_l \cdot \sin(\vf_d),
\label{equ: lambda}
\end{equation}
where $\boldsymbol{\lambda} \in \mathbb{R}^{C'}$ is a learnable channel-wise scaling vector. This formulation, grounded in Euler’s identity (\autoref{equ: euler}), effectively modulates the intermediate representation $\hat{\mX}_l$ with a phase shift parameterized by $\vf_I$.

This identity-conditioned phase modulation enables the network to impose global semantic control in a spatially adaptive manner. As shown in \autoref{fig: high-level}, the sinusoidal components allow smooth and differentiable injection of identity semantics, facilitating the generation of identity-consistent gait sequences.

By jointly incorporating high-level semantic and low-level visual conditions, our method ensures that the generated sequence $\hat{\mX}_0$ preserves appearance and motion details while maintaining strong identity consistency, thereby enhancing discriminative effectiveness.

\subsection{Training Objective}
After obtaining the identity features $\vf_I$ and $\hat{\vf}_I$, we adopt a joint loss $\gL$ to simultaneously optimize the discriminative extractor and the generative diffusion module. The overall objective is formulated as:
\begin{equation}
    \gL = \gL_{D}+\gL_{G},
\end{equation}
where $\gL_{D}$ (defined in \autoref{equ: discri-loss}) supervises $\vf_I$, while $\gL_{G} = \gL_{tri} + \gL_{ce}$ is applied to supervise the identity feature $\hat{\vf}_I$ of the generated sequence to enforce identity consistency.

\section{Experiments}
In this section, we first describe the datasets used and implementation details. We then conduct extensive experiments to evaluate \name{}, including both quantitative and qualitative analyses. Finally, comprehensive ablation studies on four datasets are performed to assess the contribution of each component within \name{}. More experiments are provided in \autoref{sec:appendixb}.

\subsection{Datasets and Evaluation Metrics}
\begin{wraptable}[10]{r}{0.53\linewidth}
\vspace{-11pt}
\centering
\small
\caption{\textbf{Implementation details}. The batch size $(P, K)$ denotes $P$ subjects and $K$ sequences per subject. The parameters $dr$, $lr$, and $wd$ refer to the decay rate, learning rate, and weight decay, respectively.}
\vspace{-10pt}
\begin{tabular}{lccc}
\toprule
Dataset & Batch Size & Optimizer & Steps \\ \midrule
SUSTech1K & (8, 4) & \multirow{4}{*}{\makecell[c]{Adam ($dr{=}0.1$)\\ $lr{=}1e\!-\!4$ \\ $wd{=}5e\!-\!4$}}& 50K\\ 
CCPG & (8, 8) & & 60K\\
GREW & (32, 2) & & 180K \\
Gait3D & (32, 2) & & 60K\\
\bottomrule
\end{tabular}
\label{tab: Statistics}
\vspace{-10pt}
\end{wraptable}

\textbf{Datasets:} We evaluate our method on four widely used datasets: SUSTech1K~\citep{SUSTech1K}, CCPG~\citep{CCPG}, GREW~\citep{GREW}, and Gait3D~\citep{Gait3D}. SUSTech1K, collected in laboratory, includes conditions such as normal, clothing changes, night, and occlusion. CCPG is designed for cross-domain evaluation, comprising four clothing-change scenarios (\ie, full-body, upper-body, lower-body, and backpacks changes). GREW and Gait3D are large-scale real-world datasets with significant challenges due to diverse environmental conditions. All training and testing splits strictly follow the official dataset protocols.

\textbf{Metrics:} Following prior work~\citep{CLTD}, we use Rank-$k$ accuracy (R-$k$) and mean Average Precision (mAP) to evaluate the performance of \name{}.

\subsection{Implementation Details}

(1) All images are resized to $64 \times 44$, and an ordered sampling strategy with a fixed sequence length of 30 frames is adopted during training. (2) We primarily employ DeepGaitV2~\citep{DeepGaitV2} as the discriminative extractor to validate \name{}, and further assess its versatility with other baselines, including GaitSet~\citep{GaitSet}, GaitGL~\citep{GaitGL}, and GaitBase~\citep{GaitBase}. (3) Dataset-specific configurations are provided in \autoref{tab: Statistics}. To ensure fairness, the batch size is halved due to the reuse of the discriminative extractor. (4) The generative diffusion module comprises convolution layers, LeakyReLU activations, batch normalization, upsampling (via linear interpolation), and spatial max pooling. Further architectural details are present in \autoref{sec:appendixa}. (5) The number of continuous frames $m$ in \autoref{equ: sample} is fixed to 5. (6) All experiments are conducted on Nvidia GeForce RTX 3090 GPUs.

\begin{table}[!t]
    \centering
    \caption{Performance comparisons on SUSTech1K. The \textbf{best} and \underline{second-best} results are highlighted in bold and underlined, respectively.}
    \vspace{-10pt}
    \resizebox{\linewidth}{!}{
    \begin{tabular}{l|l|c|cccccccc|cc}
        \toprule
        \multirow{2}{*}{Modality} & \multirow{2}{*}{Method} & \multirow{2}{*}{Venue} & \multicolumn{8}{c|}{Probe Sequence (R-1)} & \multicolumn{2}{c}{Overall}\\ \cmidrule{4-13}
         & & & NM & BG & CL & CR & UB & UN & OC & NT & R-1 & R-5 \\ \midrule
        \multirow{5}{*}{Silhouette} & GaitSet & AAAI19 & 69.1 & 68.2 & 37.4 & 65.0 & 63.1 & 61.0 & 67.2 & 23.0 & 65.0 & 84.8 \\
         & GaitPart & CVPR19 & 62.2 & 62.8 & 33.1 & 59.5 & 57.2 & 54.8 & 57.2 & 21.7 & 59.2 & 80.8 \\
         & GaitGL & ICCV21 & 67.1 & 66.2 & 35.9 & 63.3 & 61.6 & 58.1 & 66.6 & 17.9 & 63.1 & 82.8 \\
         & GaitBase & CVPR23 & 81.5 & 77.5 & \underline{49.6} & 75.8 & 75.5 & 76.7 & 81.4 & 25.9 & 76.1 & 89.4 \\
         & DeepGaitV2 & TPAMI25 & 83.5 & 79.5 & 46.3 & 76.8 & 79.1 & 78.5 & 81.1 & 27.3 & 77.4 & 90.2 \\ \midrule
        \multirow{2}{*}{\makecell[c]{Silhouette\\ + Skeleton}} & BiFusion & MTAP24 & 69.8 & 62.3 & 45.4 & 60.9 & 54.3 & 63.5 & 77.8 & 33.7 & 62.1 & 83.4 \\
         & SkeletonGait++ & AAAI24 & \underline{85.1} & \underline{82.9} & 46.6 & \underline{81.9} & \underline{80.8} & \underline{82.5} & \underline{86.2} & \textbf{47.5} & \underline{81.3} & \underline{95.5} \\\midrule
        Silhouette & \textbf{Ours} & - & \textbf{87.9} & \textbf{84.5} & \textbf{55.4} & \textbf{82.8} & \textbf{87.2} & \textbf{85.1} & \textbf{88.7} & \underline{38.6} & \textbf{83.8} & \textbf{95.8} \\
        \bottomrule
    \end{tabular}}
    \vspace{-15pt}
    \label{tab: sustech1k}
\end{table}

\subsection{Quantitative Results}

\begin{wraptable}[10]{r}{0.62\linewidth}
    \vspace{-10pt}
    \centering
    \caption{Performance comparisons on CCPG.}
    \small
    \vspace{-10pt}
    \begin{tabular}{l|c|ccccc}
        \toprule
        \multirow{2}{*}{Method} & \multirow{2}{*}{Venue} & \multicolumn{5}{c}{Gait Evaluation Protocol} \\ \cmidrule{3-7}
         & & CL & UP & DN & BG & Mean  \\ \midrule
        GaitSet & AAAI19 & 60.2 & 65.2 & 65.1 & 68.5 & 64.8 \\
        GaitPart & CVPR20 & 64.3 & 67.8 & 68.6 & 71.7 & 68.1 \\
        GaitBase & CVPR23 & 71.6 & 75.0 & 76.8 & 78.6 & 75.5 \\ 
        DeepGaitV2 & TPAMI25 & 78.6 & 84.8 & 80.7 & 89.2 & 83.3 \\ \midrule
        \textbf{Ours} & - & \textbf{80.1} & \textbf{86.9} & \textbf{81.6} & \textbf{90.9} & \textbf{84.8} \\
        \bottomrule
    \end{tabular}
    \label{tab: CCPG}
\end{wraptable}

\textbf{Evaluation on SUSTech1K and CCPG.} We compare \name{} with several recent methods~\citep{GaitSet, GaitPart, GaitGL, GaitBase, DeepGaitV2, BiFusion, SkeletonGait} on the SUSTech1K and CCPG datasets. These results underscore the superiority of \name{}. Key observations from \autoref{tab: sustech1k} are as follows: (1) Silhouette-based methods perform poorly under low-light conditions, achieving a maximum accuracy of only 27.3\%, primarily due to degraded image quality caused by insufficient lighting. Despite this, \name{} consistently outperforms these methods across all conditions, with a notable improvement of +11.3\% under the night condition compared to DeepGaitV2, which achieves the second-highest accuracy (silhouette-based methods) at 27.3\%. This highlights \name{}’s enhanced ability to extract discriminative features, especially in challenging low-quality silhouette scenarios, such as those encountered at night. (2) \name{} achieves state-of-the-art results in most conditions (seven out of eight), outperforming SkeletonGait++ (a multimodal-based method), demonstrating that our method effectively leverages silhouette data alone without relying on additional modalities. 

In \autoref{tab: CCPG}, \name{} achieves SOTA results across all scenarios, with an average Rank-1 accuracy of 84.8\%. This demonstrates that \name{} effectively combines the strengths of discriminative and generative models, significantly improving the discriminative model under various clothing conditions.

\begin{table}[!t]
    \begin{minipage}{0.45\textwidth}
        \centering
        \caption{Performance comparisons on GREW and Gait3D.}
        \vspace{-10pt}
        \resizebox{\linewidth}{!}{
            \begin{tabular}{l|c|cc|cc}
                \toprule
                \multirow{2}{*}{Method} & \multirow{2}{*}{Venue} & \multicolumn{2}{c|}{GREW} & \multicolumn{2}{c}{Gait3D} \\ \cmidrule{3-6}
                 & & Rank-1 & Rank-5 & Rank-1 & mAP \\ \midrule
                GaitSet & AAAI19 & 46.3 & 63.6 & 36.7 & 30.0 \\
                GaitPart & CVPR19 & 44.0 & 60.7 & 28.2 & 21.6 \\
                GaitGL & ICCV21 & 47.3 & 63.6 & 29.7 & 22.3 \\
                SMPLGait   & CVPR22 & -    & -    & 46.3 & 37.2 \\ 
                DANet      & CVPR23 & -    & -    & 48.0 & - \\
                GaitBase   & CVPR23 & 60.1 & -    & 64.6 & -  \\
                GaitGCI & CVPR23 & 68.5 & 80.8 & 50.3 & 39.5 \\
                HSTL   & ICCV23 & 62.7 & 76.6   & 61.3 & 55.5  \\
                DyGait    & ICCV23 & 71.4 & 83.2 & 66.3 & 56.4 \\
                QAGait & AAAI24 & 59.1 & 74.0 & 67.0 & 56.5 \\
                VPNet & CVPR24 & \underline{80.0} & \underline{89.4} & 75.4 & - \\
                CLTD & ECCV24 & 78.0 & 87.8 & 69.7 & - \\
                WaveLoss & AAAI25 & - & - & \underline{75.6} & \underline{66.5} \\
                DeepGaitV2 & TPAMI25 & 77.7 & 87.9 & 74.4 & 65.8 \\
                \textbf{Ours} & - & \textbf{81.2} & \textbf{90.8} & \textbf{78.3} & \textbf{71.2} \\
                \bottomrule
            \end{tabular}
            \label{tab: GREW}
        }
    \end{minipage}
    \hfill
    \begin{minipage}{0.53\textwidth}
        \centering
        \captionsetup{type=figure}
        \includegraphics[width=\linewidth]{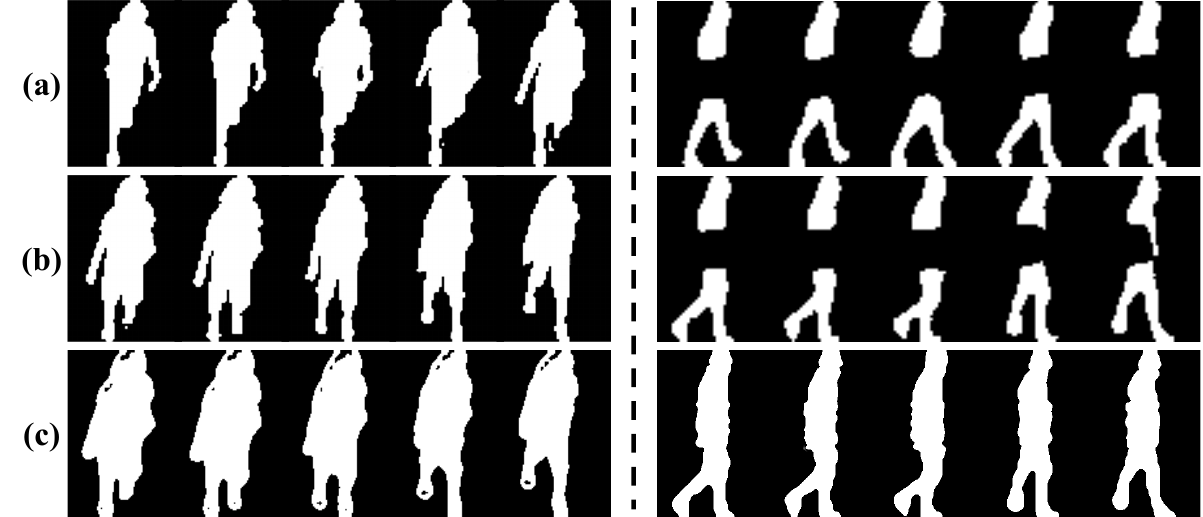}
        \caption{\textbf{Visualization of the generated sequence.} From top to bottom, sequences represent $\mX_0^m$, the ground truth of $\mX_t^n$, and $\hat{\mX}_0^n$ in \autoref{fig: framework}, respectively.}
        \label{fig: vis}
    \end{minipage}
\end{table}

\textbf{Evaluation on GREW and Gait3D.} The results on the challenging GREW and Gait3D datasets, summarized in \autoref{tab: GREW}, demonstrate that \name{} outperforms all previous methods. Specifically, on GREW, \name{} surpasses VPNet and CLTD by +1.2\% and +3.2\%, respectively, achieving a Rank-1 accuracy of 81.2\%. On Gait3D, \name{} improves upon VPNet by +2.9\% and WaveLoss~\citep{WaveLoss} by +2.7\%, reaching a Rank-1 accuracy of 78.3\%. Importantly, \name{} significantly outperforms its baseline, DeepGaitV2, with improvements of +3.5\% on GREW (81.2\% \textit{vs.} 77.7\%) and +3.9\% on Gait3D (78.3\% \textit{vs.} 74.4\%). These results further validate the effectiveness of \name{} in extracting discriminative gait features under real-world conditions.

\subsection{Qualitative Results}

\autoref{fig: vis} illustrates that the generated sequences closely resemble the original ones, demonstrating the effectiveness of our generative diffusion module in synthesizing realistic gait sequences. This success is attributed to the integration of visual details (\eg, appearance and motion) with high-level identity-aware semantic information. The visualizations also highlight the discriminative extractor's ability to learn discriminative gait features, even when the generated sequences deviate from the originals. Notably, as shown on the right side of \autoref{fig: vis}, the goal of the generative diffusion model is not merely to replicate the ground truth, but to capture and enhance discriminative gait information, thereby improving recognition robustness.

\subsection{Ablation Studies}
\label{sec: ablation}

\begin{wraptable}[14]{r}{0.64\linewidth}
    \vspace{-9pt}
    \centering
    \caption{Performance improvements (Rank-1 accuracy) of \name{} across different baselines on four datasets.}
    \vspace{-10pt}
    \small
    \begin{tabular}{c|cccc}
        \toprule
        Method & SUSTech1K & CCPG & GREW & Gait3D \\ \midrule
        GaitSet & 65.0 &64.8 & 46.3 & 36.7\\
        + \name{} & 71.3$^{\textbf{+6.3\%}}$ & 68.9$^{\textbf{+4.1\%}}$ & 54.1$^{\textbf{+7.8\%}}$ & 44.3$^{\textbf{+7.6\%}}$ \\ \midrule
        GaitGL & 63.1 & 66.2 & 47.3 & 29.7\\
        + \name{} & 69.9$^{\textbf{+6.8\%}}$ & 68.9$^{\textbf{+2.7\%}}$ & 51.5$^{\textbf{+4.2\%}}$ & 35.7$^{\textbf{+6.0\%}}$\\ \midrule
        GaitBase & 76.1 & 75.5 & 60.1 & 64.6\\
        + \name{} &84.2$^{\textbf{+8.1\%}}$ & 79.4$^{\textbf{+3.9\%}}$ & 71.1$^{\textbf{+11.0\%}}$ & 72.6$^{\textbf{+8.0\%}}$\\ \midrule
        DeepGaitV2 & 77.4 & 83.3 & 77.7 & 74.4\\
        + \name{} & 83.8$^{\textbf{+6.4\%}}$ & 84.8$^{\textbf{+1.5\%}}$ & 81.2$^{\textbf{+3.5\%}}$ & 78.3$^{\textbf{+3.9\%}}$\\ 
        \bottomrule
    \end{tabular}
    \label{tab: versatility}
\end{wraptable}

\textbf{Versatility of \name{}.} \autoref{tab: versatility} demonstrates that our method significantly improves performance across four discriminative extractors on four datasets, highlighting the effectiveness and versatility of collaboratively integrating discriminative and generative diffusion models for gait recognition. Notably, we observe that incorporating \name{} with non-temporal modeling methods (\eg, GaitSet and GaitBase) yields greater performance improvements compared to temporal modeling methods (\eg, GaitGL, and DeepGaitV2). This is due to the generative diffusion model's ability to introduce rich temporal dynamics, which particularly benefits non-temporal modeling methods.

\begin{table}[!h]
    \centering
    \caption{The ablation study of \multicondition{} strategy.}
    \small
    \vspace{-10pt}
    \begin{tabular}{cc|cccc}
        \toprule
        High-level & Low-level & SUSTech1K & CCPG & GREW & Gait3D \\ \midrule
        \XSolidBrush & \XSolidBrush & 77.4 & 83.3 & 77.7 & 74.4 \\
        \Checkmark & \XSolidBrush & 81.9 & 84.0 & 80.4 & 77.4 \\
        \XSolidBrush & \Checkmark & 81.3 & 83.7 & 79.9& 77.2 \\
        \Checkmark & \Checkmark &\textbf{83.8} & \textbf{84.8} & \textbf{81.2} & \textbf{78.3} \\
        \bottomrule
    \end{tabular}
    \label{tab: ablation_multi}
\end{table}

\textbf{Effectiveness of \multicondition{} strategy.}
\autoref{tab: ablation_multi} investigates the impact of the \multicondition{} strategy. The results indicate that both conditions independently improve recognition accuracy, confirming that the extractor $\gD$ effectively learns discriminative identity features. Moreover, the combination of both conditions leads to even greater performance, underscoring their complementary properties.

\begin{table}
    \centering
    \caption{The ablation study of \highcontrol{}.}
    \vspace{-10pt}
    \small
    \begin{tabular}{c|cccc}
        \toprule
        Method & SUSTech1K & CCPG & GREW & Gait3D \\ \midrule
        Baseline & 81.3 & 83.7 & 79.9 & 77.2 \\
        \textit{w/} addition & 83.0 & 84.4 & 80.7 & 77.6 \\
        \textit{w/} Ours & \textbf{83.8} & \textbf{84.8} & \textbf{81.2} & \textbf{78.3} \\
        \bottomrule
    \end{tabular}
    \label{tab: ablation_highlevel}
\end{table}

\textbf{Effectiveness of \highcontrol{}.} \autoref{tab: ablation_highlevel} evaluates the effectiveness of \highcontrol{} by comparing it with the baseline (low-level conditional control only) and an element-wise addition strategy. The results demonstrate that our control strategy outperforms direct element-wise addition, highlighting the advantages of \highcontrol{} in improving the identity consistency and quality of generated sequences.

\begin{wraptable}[9]{r}{0.48\linewidth}
    \vspace{-10pt}
    \centering
    \caption{The ablation study of the number of continuous frames $m$ in \autoref{equ: sample}.}
    \small
    \vspace{-10pt}
    \begin{tabular}{c|cccc}
        \toprule
        $m$ & SUSTech1K & CCPG & GREW & Gait3D \\ \midrule
        1 & 81.2 & 83.7 & 79.6 & 76.9 \\
        3 & 83.3 & 84.5 & 80.5 & 77.6 \\
        5 & 83.8 & \textbf{84.8} & \textbf{81.2} & \textbf{78.3} \\
        7 & \textbf{84.0} & \textbf{84.8} & 80.9 & 78.0 \\
        9 & 83.4 & 84.2 & 80.6 & 77.8 \\
        \bottomrule
    \end{tabular}
    \label{tab: ablation_l}
\end{wraptable}

\textbf{Impact of the number of continuous frames $m$.} The number of continuous frames, $m$, plays a crucial role in balancing the low-level visual and high-level semantic conditions. As listed in \autoref{tab: ablation_l}, experiments with different values of $m$ (\ie, $m \in \{1,3,5,7,9\}$) reveal the following trends: (1) When low-level visual information is highly limited (\ie, $m=1$), the generative diffusion model struggles to learn discriminative features, resulting in suboptimal performance compared to using a single condition. (2) When $m$ is too large, the generative diffusion model does not need to learn sufficiently, limiting its ability to guide the discriminative extractor and degrading performance. Based on a comprehensive evaluation across four datasets, we set $m$ to 5.

\begin{table}[h]
    \centering
    \caption{The ablation study of the learnable vector $\boldsymbol{\lambda}$ in \autoref{equ: lambda}.}
    \small
    \vspace{-10pt}
    \begin{tabular}{c|cccc}
        \toprule
        $\boldsymbol{\lambda}$ & SUSTech1K & CCPG & GREW & Gait3D \\ \midrule
        1 & 83.2 & 84.3 & 80.3 & 77.4 \\
        learnable scalar & 83.5 & 84.4 & 80.8 & 77.8 \\
        \textbf{learnable vector} & \textbf{83.8} & \textbf{84.8} & \textbf{81.2} & \textbf{78.3} \\
        \bottomrule
    \end{tabular}
    \label{tab: ablation_lambda}
\end{table}

\textbf{Impact of the learnable vector $\boldsymbol{\lambda}$.}
\autoref{tab: ablation_lambda} compares different strategies for $\boldsymbol{\lambda}$ (\ie, a fixed value, a learnable scalar, and our learnable vector) in \autoref{equ: lambda}. The results show that using a learnable weight to control the adjustment intensity generally enhances recognition performance. Furthermore, the learnable vector achieves the best results, as it allows for adaptive adjustments across different channels.

\begin{wraptable}[15]{r}{0.5\linewidth}
    \centering
    \vspace{-10pt}
    \caption{Training and testing resource consumption on Gait3D. Training is calculated across four GPUs, while testing uses a single GPU.}
    \vspace{-10pt}
    \small
    \begin{tabular}{c|c|c}
        \toprule
        Method & Training (hour) & Testing (second)\\ \midrule
        GaitSet & 0.95 & 41 \\
        + \name{} & 1.08 (\textbf{+13.7\%}) & 41 (\textbf{+0\%}) \\ \midrule
        GaitGL & 2.91 & 43\\
        + \name{} & 3.35 (\textbf{+15.1\%}) & 43 (\textbf{+0\%}) \\ \midrule
        GaitBase & 5.83 & 77\\
        + \name{} & 6.29 (\textbf{+7.9\%}) & 77 (\textbf{+0\%}) \\ \midrule
        DeepGaitV2 & 9.98 & 95\\
        + \name{} & 10.94 (\textbf{+9.6\%}) & 95 (\textbf{+0\%}) \\
        \bottomrule
    \end{tabular}
    \label{tab: ablation_time}
\end{wraptable}

\textbf{Training and Testing Resource Consumption.} 
\autoref{tab: ablation_time} analyzes the resource consumption of our method on Gait3D during both training and testing. While training demands increase by 7.9\% to 15.1\% compared to baselines (\ie, GaitSet, GaitGL, GaitBase, and DeepGaitV2), the overhead remains acceptable by halving the batch size, despite the reuse of the discriminative extractor and the introduction of the generative diffusion model. Notably, testing resource requirements remain unchanged. The results in \autoref{tab: versatility} and \autoref{tab: ablation_time} demonstrate that \name{} achieves significant performance gains without compromising testing efficiency.

\section{Conclusion}
In this paper, we propose \name{}, an novel gait recognition framework that collaboratively combines the data distribution modeling capabilities of diffusion models with the semantic representation learning strengths of discriminative models. We introduce a \multicondition{} strategy that integrates high-level identity-aware semantic information with low-level visual details to guide the generation process. Furthermore, ensuring identity consistency of generated sequences enhances the discriminative model's ability to learn robust gait features. We assess the effectiveness and versatility of \name{} on four datasets.

\subsubsection*{Acknowledgments}
This work is supported by National Natural Science Foundation of China (No. 62376102).

\bibliography{main}

@inproceedings{WaveLoss,
  title={WaveLoss: An Adaptive Dynamic Loss for Deep Gait Recognition},
  author={Wang, Zicheng and Wu, Qiuxia},
  booktitle={Proceedings of the AAAI Conference on Artificial Intelligence},
  volume={39},
  number={8},
  pages={8259--8267},
  year={2025}
}

@inproceedings{GaitGS,
  title={Gaitgs: Temporal feature learning in granularity and span dimension for gait recognition},
  author={Xiong, Haijun and Deng, Yunze and Feng, Bin and Wang, Xinggang and Liu, Wenyu},
  booktitle={2024 IEEE International Conference on Image Processing (ICIP)},
  pages={2410--2416},
  year={2024},
  organization={IEEE}
}

@inproceedings{CCPG,
  title={An in-depth exploration of person re-identification and gait recognition in cloth-changing conditions},
  author={Li, Weijia and Hou, Saihui and Zhang, Chunjie and Cao, Chunshui and Liu, Xu and Huang, Yongzhen and Zhao, Yao},
  booktitle={Proceedings of the IEEE/CVF Conference on Computer Vision and Pattern Recognition},
  pages={13824--13833},
  year={2023}
}

@article{GaitTR,
  title={Spatial transformer network on skeleton-based gait recognition},
  author={Zhang, Cun and Chen, Xing-Peng and Han, Guo-Qiang and Liu, Xiang-Jie},
  journal={Expert Systems},
  volume={40},
  number={6},
  pages={e13244},
  year={2023},
  publisher={Wiley Online Library}
}

@inproceedings{GaitMixer,
  title={Gaitmixer: skeleton-based gait representation learning via wide-spectrum multi-axial mixer},
  author={Pinyoanuntapong, Ekkasit and Ali, Ayman and Wang, Pu and Lee, Minwoo and Chen, Chen},
  booktitle={ICASSP 2023-2023 IEEE International Conference on Acoustics, Speech and Signal Processing (ICASSP)},
  pages={1--5},
  year={2023},
  organization={IEEE}
}

@inproceedings{DenoisingGait,
  title={On Denoising Walking Videos for Gait Recognition},
  author={Jin, Dongyang and Fan, Chao and Ma, Jingzhe and Zhou, Jingkai and Chen, Weihua and Yu, Shiqi},
  booktitle={Proceedings of the IEEE/CVF Conference on Computer Vision and Pattern Recognition},
  pages={12347--12357},
  year={2025}
}

@article{4Real,
  title={4real: Towards photorealistic 4d scene generation via video diffusion models},
  author={Yu, Heng and Wang, Chaoyang and Zhuang, Peiye and Menapace, Willi and Siarohin, Aliaksandr and Cao, Junli and Jeni, L{\'a}szl{\'o} and Tulyakov, Sergey and Lee, Hsin-Ying},
  journal={Advances in Neural Information Processing Systems},
  volume={37},
  pages={45256--45280},
  year={2024}
}

@inproceedings{CAT4D,
  title={Cat4d: Create anything in 4d with multi-view video diffusion models},
  author={Wu, Rundi and Gao, Ruiqi and Poole, Ben and Trevithick, Alex and Zheng, Changxi and Barron, Jonathan T and Holynski, Aleksander},
  booktitle={Proceedings of the IEEE/CVF Conference on Computer Vision and Pattern Recognition},
  pages={26057--26068},
  year={2025}
}

@article{venkat2011robust,
  title={Robust gait recognition by learning and exploiting sub-gait characteristics},
  author={Venkat, Ibrahim and De Wilde, Philippe},
  journal={International Journal of Computer Vision},
  volume={91},
  number={1},
  pages={7--23},
  year={2011},
  publisher={Springer}
}

@article{MambaGait,
  title={MambaGait: Gait recognition approach combining explicit representation and implicit state space model},
  author={Xiong, Haijun and Feng, Bin and Wang, Bang and Wang, Xinggang and Liu, Wenyu},
  journal={Image and Vision Computing},
  pages={105597},
  year={2025},
  publisher={Elsevier}
}

@article{sepas2022deep,
  title={Deep Gait Recognition: A Survey},
  author={Sepas-Moghaddam, Alireza and Etemad, Ali},
  journal = {IEEE transactions on pattern analysis and machine intelligence},
  volume={45},
  number={1},
  pages={264--284},
  year={2022}
}

@inproceedings{GaitSet,
  title={Gaitset: Regarding gait as a set for cross-view gait recognition},
  author={Chao, Hanqing and He, Yiwei and Zhang, Junping and Feng, Jianfeng},
  booktitle={Proceedings of the AAAI Conference on Artificial Intelligence},
  volume={33},
  number={01},
  pages={8126--8133},
  year={2019}
}

@inproceedings{GaitBase,
  title={Opengait: Revisiting gait recognition towards better practicality},
  author={Fan, Chao and Liang, Junhao and Shen, Chuanfu and Hou, Saihui and Huang, Yongzhen and Yu, Shiqi},
  booktitle={Proceedings of the IEEE/CVF Conference on Computer Vision and Pattern Recognition},
  pages={9707--9716},
  year={2023}
}

@article{DDPM,
  title={Denoising Diffusion Probabilistic Models},
  author={Ho, Jonathan and Jain, Ajay and Abbeel, Pieter},
  journal={Advances in Neural Information Processing Systems},
  volume={33},
  pages={6840--6851},
  year={2020}
}

@inproceedings{HSTL,
  title={Hierarchical Spatio-Temporal Representation Learning for Gait Recognition},
  author={Wang, Lei and Liu, Bo and Liang, Fangfang and Wang, Bincheng},
  booktitle={Proceedings of the IEEE/CVF International Conference on Computer Vision},
  pages={19639--19649},
  year={2023}
}

@inproceedings{DyGait,
    author={Wang, Ming and Guo, Xianda and Lin, Beibei and Yang, Tian and Zhu, Zheng and Li, Lincheng and Zhang, Shunli and Yu, Xin},
    title={DyGait: Exploiting Dynamic Representations for High-performance Gait Recognition},
    booktitle={Proceedings of the IEEE/CVF International Conference on Computer Vision},
    year={2023},
    pages={13424-13433}
}

@article{VideoDM,
  title={Video Diffusion Models},
  author={Ho, Jonathan and Salimans, Tim and Gritsenko, Alexey and Chan, William and Norouzi, Mohammad and Fleet, David J},
  journal={Advances in Neural Information Processing Systems},
  volume={35},
  pages={8633--8646},
  year={2022}
}

@inproceedings{DDIM,
  title={Denoising Diffusion Implicit Models},
  author={Song, Jiaming and Meng, Chenlin and Ermon, Stefano},
  booktitle={International Conference on Learning Representations},
  year={2021}
}

@inproceedings{DiffPose,
  title={DiffPose: SpatioTemporal diffusion model for video-based human pose estimation},
  author={Feng, Runyang and Gao, Yixing and Tse, Tze Ho Elden and Ma, Xueqing and Chang, Hyung Jin},
  booktitle={Proceedings of the IEEE/CVF International Conference on Computer Vision},
  pages={14861--14872},
  year={2023}
}

@inproceedings{MacDiff,
  title={MacDiff: Unified Skeleton Modeling with Masked Conditional Diffusion},
  author={Lehong Wu and Lilang Lin and Jiahang Zhang and Yiyang Ma and Jiaying Liu},
  booktitle={European Conference on Computer Vision},
  pages={110--128},
  year={2024},
  organization={Springer}
}

@inproceedings{FinePose,
  title={FinePOSE: Fine-Grained Prompt-Driven 3D Human Pose Estimation via Diffusion Models},
  author={Xu, Jinglin and Guo, Yijie and Peng, Yuxin},
  booktitle={Proceedings of the IEEE/CVF Conference on Computer Vision and Pattern Recognition},
  pages={561--570},
  year={2024}
}

@inproceedings{DPMesh,
  title={DPMesh: Exploiting Diffusion Prior for Occluded Human Mesh Recovery},
  author={Zhu, Yixuan and Li, Ao and Tang, Yansong and Zhao, Wenliang and Zhou, Jie and Lu, Jiwen},
  booktitle={Proceedings of the IEEE/CVF Conference on Computer Vision and Pattern Recognition},
  pages={1101--1110},
  year={2024}
}

@inproceedings{GREW,
  title={Gait recognition in the wild: A benchmark},
  author={Zhu, Zheng and Guo, Xianda and Yang, Tian and Huang, Junjie and Deng, Jiankang and Huang, Guan and Du, Dalong and Lu, Jiwen and Zhou, Jie},
  booktitle={Proceedings of the IEEE/CVF International Conference on Computer Vision},
  pages={14789--14799},
  year={2021}
}

@inproceedings{GaitGraph,
  title={Gaitgraph: Graph convolutional network for skeleton-based gait recognition},
  author={Teepe, Torben and Khan, Ali and Gilg, Johannes and Herzog, Fabian and H{\"o}rmann, Stefan and Rigoll, Gerhard},
  booktitle={2021 IEEE international conference on image processing (ICIP)},
  pages={2314--2318},
  year={2021},
  organization={IEEE}
}

@inproceedings{Gait3D,
  title={Gait recognition in the wild with dense 3d representations and a benchmark},
  author={Zheng, Jinkai and Liu, Xinchen and Liu, Wu and He, Lingxiao and Yan, Chenggang and Mei, Tao},
  booktitle={Proceedings of the IEEE/CVF Conference on Computer Vision and Pattern Recognition},
  pages={20228--20237},
  year={2022}
}

@inproceedings{GaitGraph2,
  title={Towards a deeper understanding of skeleton-based gait recognition},
  author={Teepe, Torben and Gilg, Johannes and Herzog, Fabian and H{\"o}rmann, Stefan and Rigoll, Gerhard},
  booktitle={Proceedings of the IEEE/CVF Conference on Computer Vision and Pattern Recognition (CVPR) Workshops},
  pages={1569--1577},
  year={2022}
}

@inproceedings{GPGait,
  title={Gpgait: Generalized pose-based gait recognition},
  author={Fu, Yang and Meng, Shibei and Hou, Saihui and Hu, Xuecai and Huang, Yongzhen},
  booktitle={Proceedings of the IEEE/CVF International Conference on Computer Vision},
  pages={19595--19604},
  year={2023}
}

@article{CycleGait,
  title={A strong and robust skeleton-based gait recognition method with gait periodicity priors},
  author={Li, Na and Zhao, Xinbo},
  journal={IEEE Transactions on Multimedia},
  volume={25},
  pages={3046--3058},
  year={2022}
}

@article{PoseGait,
  title={A model-based gait recognition method with body pose and human prior knowledge},
  author={Liao, Rijun and Yu, Shiqi and An, Weizhi and Huang, Yongzhen},
  journal={Pattern Recognition},
  volume={98},
  pages={107069},
  year={2020}
}

@article{CAG,
  title={Condition-adaptive graph convolution learning for skeleton-based gait recognition},
  author={Huang, Xiaohu and Wang, Xinggang and Jin, Zhidianqiu and Yang, Bo and He, Botao and Feng, Bin and Liu, Wenyu},
  journal={IEEE Transactions on Image Processing},
  volume={32},
  pages={4773--4784},
  year={2023}
}

@inproceedings{GaitPart,
  title={Gaitpart: Temporal part-based model for gait recognition},
  author={Fan, Chao and Peng, Yunjie and Cao, Chunshui and Liu, Xu and Hou, Saihui and Chi, Jiannan and Huang, Yongzhen and Li, Qing and He, Zhiqiang},
  booktitle={Proceedings of the IEEE/CVF Conference on Computer Vision and Pattern Recognition},
  pages={14225--14233},
  year={2020}
}

@inproceedings{GaitGL,
  title={Gait recognition via effective global-local feature representation and local temporal aggregation},
  author={Lin, Beibei and Zhang, Shunli and Yu, Xin},
  booktitle={Proceedings of the IEEE/CVF International Conference on Computer Vision},
  pages={14648--14656},
  year={2021}
}

@inproceedings{QAGait,
  title={QAGait: Revisit Gait Recognition from a Quality Perspective},
  author={Wang, Zengbin and Hou, Saihui and Zhang, Man and Liu, Xu and Cao, Chunshui and Huang, Yongzhen and Li, Peipei and Xu, Shibiao},
  booktitle={Proceedings of the AAAI Conference on Artificial Intelligence},
  volume={38},
  number={6},
  pages={5785--5793},
  year={2024}
}

@inproceedings{DANet,
  title={Dynamic aggregated network for gait recognition},
  author={Ma, Kang and Fu, Ying and Zheng, Dezhi and Cao, Chunshui and Hu, Xuecai and Huang, Yongzhen},
  booktitle={Proceedings of the IEEE/CVF Conference on Computer Vision and Pattern Recognition},
  pages={22076--22085},
  year={2023}
}

@inproceedings{CSTL,
  title={Context-sensitive temporal feature learning for gait recognition},
  author={Huang, Xiaohu and Zhu, Duowang and Wang, Hao and Wang, Xinggang and Yang, Bo and He, Botao and Liu, Wenyu and Feng, Bin},
  booktitle={Proceedings of the IEEE/CVF International Conference on Computer Vision},
  pages={12909--12918},
  year={2021}
}

@inproceedings{VPNet,
  title={Learning Visual Prompt for Gait Recognition},
  author={Ma, Kang and Fu, Ying and Cao, Chunshui and Hou, Saihui and Huang, Yongzhen and Zheng, Dezhi},
  booktitle={Proceedings of the IEEE/CVF Conference on Computer Vision and Pattern Recognition},
  pages={593--603},
  year={2024}
}

@article{DeepGaitV2,
  title={OpenGait: A Comprehensive Benchmark Study for Gait Recognition towards Better Practicality},
  author={Fan, Chao and Hou, Saihui and Liang, Junhao and Shen, Chuanfu and Ma, Jingzhe and Jin, Dongyang and Huang, Yongzhen and Yu, Shiqi},
  journal={IEEE Transactions on Pattern Analysis and Machine Intelligence},
  volume={47},
  number={10},
  pages={8397--8414},
  year={2025}
}

@inproceedings{GaitGCI,
  title={Gaitgci: Generative counterfactual intervention for gait recognition},
  author={Dou, Huanzhang and Zhang, Pengyi and Su, Wei and Yu, Yunlong and Lin, Yining and Li, Xi},
  booktitle={Proceedings of the IEEE/CVF Conference on Computer Vision and Pattern Recognition},
  pages={5578--5588},
  year={2023}
}

@inproceedings{CLTD,
  title={Causality-inspired Discriminative Feature Learning in Triple Domains for Gait Recognition},
  author={Xiong, Haijun and Feng, Bin and Wang, Xinggang and Liu, Wenyu},
  booktitle={European Conference on Computer Vision},
  pages={251--270},
  year={2024},
  organization={Springer}
}

@inproceedings{GaitCSV,
  title={Causal intervention for sparse-view gait recognition},
  author={Wang, Jilong and Hou, Saihui and Huang, Yan and Cao, Chunshui and Liu, Xu and Huang, Yongzhen and Wang, Liang},
  booktitle={Proceedings of the 31st ACM International Conference on Multimedia},
  pages={77--85},
  year={2023}
}

@inproceedings{SUSTech1K,
  title={Lidargait: Benchmarking 3d gait recognition with point clouds},
  author={Shen, Chuanfu and Fan, Chao and Wu, Wei and Wang, Rui and Huang, George Q and Yu, Shiqi},
  booktitle={Proceedings of the IEEE/CVF Conference on Computer Vision and Pattern Recognition},
  pages={1054--1063},
  year={2023}
}

@inproceedings{BigGait,
  title={BigGait: Learning Gait Representation You Want by Large Vision Models},
  author={Ye, Dingqiang and Fan, Chao and Ma, Jingzhe and Liu, Xiaoming and Yu, Shiqi},
  booktitle={Proceedings of the IEEE/CVF Conference on Computer Vision and Pattern Recognition},
  pages={200--210},
  year={2024}
}

@inproceedings{HumanMac,
  title={Humanmac: Masked motion completion for human motion prediction},
  author={Chen, Ling-Hao and Zhang, Jiawei and Li, Yewen and Pang, Yiren and Xia, Xiaobo and Liu, Tongliang},
  booktitle={Proceedings of the IEEE/CVF Conference on Computer Vision and Pattern Recognition},
  pages={9544--9555},
  year={2023}
}

@inproceedings{HOIAnimator,
  title={HOIAnimator: Generating Text-prompt Human-object Animations using Novel Perceptive Diffusion Models},
  author={Song, Wenfeng and Zhang, Xinyu and Li, Shuai and Gao, Yang and Hao, Aimin and Hou, Xia and Chen, Chenglizhao and Li, Ning and Qin, Hong},
  booktitle={Proceedings of the IEEE/CVF Conference on Computer Vision and Pattern Recognition},
  pages={811--820},
  year={2024}
}

@inproceedings{ControlNet,
  title={Adding conditional control to text-to-image diffusion models},
  author={Zhang, Lvmin and Rao, Anyi and Agrawala, Maneesh},
  booktitle={Proceedings of the IEEE/CVF International Conference on Computer Vision},
  pages={3836--3847},
  year={2023}
}

@inproceedings{P2P-Bridge,
  title={P2P-Bridge: Diffusion Bridges for 3D Point Cloud Denoising},
  author={Vogel, Mathias and Tateno, Keisuke and Pollefeys, Marc and Tombari, Federico and Rakotosaona, Marie-Julie and Engelmann, Francis},
  booktitle={European Conference on Computer Vision},
  pages={184--201},
  year={2024},
  organization={Springer}
}

@inproceedings{SatSynth,
  title={Satsynth: Augmenting image-mask pairs through diffusion models for aerial semantic segmentation},
  author={Toker, Aysim and Eisenberger, Marvin and Cremers, Daniel and Leal-Taix{\'e}, Laura},
  booktitle={Proceedings of the IEEE/CVF Conference on Computer Vision and Pattern Recognition},
  pages={27695--27705},
  year={2024}
}

@inproceedings{RAVE,
  title={Rave: Randomized noise shuffling for fast and consistent video editing with diffusion models},
  author={Kara, Ozgur and Kurtkaya, Bariscan and Yesiltepe, Hidir and Rehg, James M and Yanardag, Pinar},
  booktitle={Proceedings of the IEEE/CVF Conference on Computer Vision and Pattern Recognition},
  pages={6507--6516},
  year={2024}
}

@inproceedings{FreeDiff,
  title={FreeDiff: Progressive Frequency Truncation for Image Editing with Diffusion Models},
  author={Wu, Wei and Fan, Qingnan and Qin, Shuai and Gu, Hong and Zhao, Ruoyu and Chan, Antoni B},
  booktitle={European Conference on Computer Vision},
  pages={194--209},
  year={2024},
  organization={Springer}
}

@article{LAMP,
  title={LAMP: Learn A Motion Pattern for Few-Shot-Based Video Generation},
  author={Wu, Ruiqi and Chen, Liangyu and Yang, Tong and Guo, Chunle and Li, Chongyi and Zhang, Xiangyu},
  journal={arXiv preprint arXiv:2310.10769},
  year={2023}
}

@inproceedings{SkeletonGait,
  title={SkeletonGait: Gait Recognition Using Skeleton Maps},
  author={Fan, Chao and Ma, Jingzhe and Jin, Dongyang and Shen, Chuanfu and Yu, Shiqi},
  booktitle={Proceedings of the AAAI Conference on Artificial Intelligence},
  volume={38},
  number={2},
  pages={1662--1669},
  year={2024}
}

@inproceedings{I2v-adapter,
  title={I2v-adapter: A general image-to-video adapter for diffusion models},
  author={Guo, Xun and Zheng, Mingwu and Hou, Liang and Gao, Yuan and Deng, Yufan and Wan, Pengfei and Zhang, Di and Liu, Yufan and Hu, Weiming and Zha, Zhengjun and others},
  booktitle={ACM SIGGRAPH 2024 Conference Papers},
  pages={1--12},
  year={2024}
}

@inproceedings{AYG,
  title={Align your gaussians: Text-to-4d with dynamic 3d gaussians and composed diffusion models},
  author={Ling, Huan and Kim, Seung Wook and Torralba, Antonio and Fidler, Sanja and Kreis, Karsten},
  booktitle={Proceedings of the IEEE/CVF Conference on Computer Vision and Pattern Recognition},
  pages={8576--8588},
  year={2024}
}

@inproceedings{MTSGait,
  title={Gait recognition in the wild with multi-hop temporal switch},
  author={Zheng, Jinkai and Liu, Xinchen and Gu, Xiaoyan and Sun, Yaoqi and Gan, Chuang and Zhang, Jiyong and Liu, Wu and Yan, Chenggang},
  booktitle={Proceedings of the 30th ACM International Conference on Multimedia},
  pages={6136--6145},
  year={2022}
}

@article{BiFusion,
  title={Learning rich features for gait recognition by integrating skeletons and silhouettes},
  author={Peng, Yunjie and Ma, Kang and Zhang, Yang and He, Zhiqiang},
  journal={MTAP},
  volume={83},
  number={3},
  pages={7273--7294},
  year={2024},
  publisher={Springer}
}

@article{OU-MVLP,
  title={Multi-view large population gait dataset and its performance evaluation for cross-view gait recognition},
  author={Takemura, Noriko and Makihara, Yasushi and Muramatsu, Daigo and Echigo, Tomio and Yagi, Yasushi},
  journal={IPSJ transactions on Computer Vision and Applications},
  volume={10},
  pages={1--14},
  year={2018},
  publisher={Springer}
}

@inproceedings{CASIA-B,
  title={A framework for evaluating the effect of view angle, clothing and carrying condition on gait recognition},
  author={Yu, Shiqi and Tan, Daoliang and Tan, Tieniu},
  booktitle={18th international conference on pattern recognition (ICPR'06)},
  volume={4},
  pages={441--444},
  year={2006},
  organization={IEEE}
}

@inproceedings{GLGait,
  title={Glgait: a global-local temporal receptive field network for gait recognition in the wild},
  author={Peng, Guozhen and Wang, Yunhong and Zhao, Yuwei and Zhang, Shaoxiong and Li, Annan},
  booktitle={Proceedings of the 32nd ACM International Conference on Multimedia},
  pages={826--835},
  year={2024}
}

@inproceedings{XGait,
  title={It takes two: Accurate gait recognition in the wild via cross-granularity alignment},
  author={Zheng, Jinkai and Liu, Xinchen and Zhang, Boyue and Yan, Chenggang and Zhang, Jiyong and Liu, Wu and Zhang, Yongdong},
  booktitle={Proceedings of the 32nd ACM International Conference on Multimedia},
  pages={8786--8794},
  year={2024}
}

@inproceedings{ParsingGait,
  title={Parsing is all you need for accurate gait recognition in the wild},
  author={Zheng, Jinkai and Liu, Xinchen and Wang, Shuai and Wang, Lihao and Yan, Chenggang and Liu, Wu},
  booktitle={Proceedings of the 31st ACM International Conference on Multimedia},
  pages={116--124},
  year={2023}
}

@inproceedings{DD-GCN,
  title={Dd-gcn: Directed diffusion graph convolutional network for skeleton-based human action recognition},
  author={Li, Chang and Huang, Qian and Mao, Yingchi},
  booktitle={2023 IEEE International Conference on Multimedia and Expo (ICME)},
  pages={786--791},
  year={2023},
  organization={IEEE}
}

@inproceedings{HMDiff,
  title={Distribution-aligned diffusion for human mesh recovery},
  author={Foo, Lin Geng and Gong, Jia and Rahmani, Hossein and Liu, Jun},
  booktitle={Proceedings of the IEEE/CVF International Conference on Computer Vision},
  pages={9221--9232},
  year={2023}
}
\bibliographystyle{iclr2026_conference}

\appendix
\section{Architecture of Generative Diffusion Module}
\label{sec:appendixa}
In \autoref{tab: architecture}, we present the architectural details of the generative module $\gG$. Each block consists of two convolutional layers, accompanied by batch normalization and LeakyReLU activation.

\begin{table}[h]
    \begin{minipage}[t]{0.45\linewidth}
        \small
        \centering
        \caption{Architecture of the generative diffusion module consists of Blocks. Further, HCM represents the \highcontrol{}.}
        \vspace{-10pt}
        \begin{tabular}{c|c|c}
            \toprule
            \textbf{Module} & \textbf{Output size} & \textbf{Kernel size} \\ \midrule
            \multirow{2}{*}{Block} & \multirow{2}{*}{(32, $T$, 64, 44)} & (7, 5, 5)\\
             & & (5, 3, 3) \\ \midrule
            Pooling & (32, $T$, 32, 22) & (1, 2, 2)\\ \midrule
            HCM & (32, $T$, 32, 22) & - \\ \midrule
            \multirow{2}{*}{Block} & \multirow{2}{*}{(64, $T$, 32, 22)} & (5, 3, 3)\\
             & & (3, 3, 3) \\ \midrule
            Pooling & (64, $T$, 16, 11) & (1, 2, 2)\\ \midrule
            HCM & (64, $T$, 16, 11) & - \\ \midrule
            \multirow{2}{*}{Block} & \multirow{2}{*}{(128, $T$, 16, 11)} & (3, 3, 3)\\
             & & (3, 3, 3) \\ \midrule
            HCM & (128, $T$, 16, 11) & - \\ \midrule
            \multirow{2}{*}{Block} & \multirow{2}{*}{(64, $T$, 16, 11)} & (3, 3, 3)\\
             & & (3, 3, 3)\\ \midrule
            HCM & (64, $T$, 16, 11) & - \\ \midrule
            UpSample & (64, $T$, 32, 22) & - \\ \midrule
            \multirow{2}{*}{Block} & \multirow{2}{*}{(32, $T$, 32, 22)} & (3, 3, 3) \\
             & & (5, 3, 3) \\ \midrule
            HCM & (32, $T$, 32, 22) & - \\ \midrule
            UpSample & (32, $T$, 64, 44) & - \\ \midrule
            \multirow{2}{*}{Block} & \multirow{2}{*}{(1, $T$, 64, 44)} & (5, 3, 3) \\
             & & (7, 5, 5)\\ \midrule
             Norm & (1, $T$, 64, 44) & - \\
            \bottomrule
        \end{tabular}
        \label{tab: architecture}
    \end{minipage}
    \hfill
    \begin{minipage}[t]{0.53\linewidth}
        \centering
        \caption{Performance improvements on CASIA-B and OU-MVLP datasets.}
        \small
        \vspace{-10pt}
        \begin{tabular}{c|cc}
        \toprule
            Method & CASIA-B & OU-MVLP \\ \midrule
            DeepGaitV2 & 89.6\% & 91.9\%  \\
            + \name{} & 89.9\% (\textbf{+0.3\%}) & 92.1 (\textbf{+0.2\%}) \\
            \bottomrule
        \end{tabular}
        \label{tab: CASIA-B and OUMVLP}
    
        \vspace{15pt}
        
        \caption{Performance improvements (Rank-1 accuracy) of \name{} across GaitPart on four datasets.}
        \vspace{-10pt}
        \resizebox{\linewidth}{!}{
        \begin{tabular}{c|cccc}
            \toprule
            Method & SUSTech1K & CCPG & GREW & Gait3D \\ \midrule
            GaitPart & 59.2 & 68.1 & 44.0 & 28.2\\
            + \name{} & 65.3$^{\textbf{+6.1\%}}$ & 70.4$^{\textbf{+2.3\%}}$ & 49.2$^{\textbf{+5.2\%}}$ & 34.9$^{\textbf{+6.7\%}}$ \\
            \bottomrule
        \end{tabular}}
        \label{tab: Gaitpart}
        \end{minipage}
\end{table}

\section{Experiments}
\label{sec:appendixb}
\subsection{Evaluation on CASIA-B and OU-MVLP}
As shown in \autoref{tab: CASIA-B and OUMVLP}, our method achieves 89.9\% and 92.1\% on CASIA-B and OUMVLP datasets~\citep{CASIA-B, OU-MVLP}, respectively, with improvements of 0.3\% and 0.2\% to DeepGaitV2~\citep{DeepGaitV2}, proving its effectiveness.

\subsection{GaitPart across CoD\texorpdfstring{$^2$}{²}}
As shown in \autoref{tab: Gaitpart}, we integrate our \name{} with GaitPart~\citep{GaitPart} to assess the versatility. 

\end{document}